\documentclass[conference]{IEEEtran}
\IEEEoverridecommandlockouts
\usepackage{cite}
\usepackage{amsmath,amssymb,amsfonts}
\usepackage{algorithmic}

\usepackage[ruled,vlined]{algorithm2e}
\usepackage{siunitx}
\SetArgSty{}

\usepackage{graphicx}
\usepackage{textcomp}
\usepackage{xcolor}
\def\BibTeX{{\rm B\kern-.05em{\sc i\kern-.025em b}\kern-.08em
    T\kern-.1667em\lower.7ex\hbox{E}\kern-.125emX}}
\begin{document}

\title{\textbf{\textit{Evaluation of Hyperparameter-Optimization Approaches in an Industrial Federated Learning System}} \huge \\ 
}
\author{\IEEEauthorblockN{Stephanie Holly}
\IEEEauthorblockA{Siemens Technology and\\ TU Wien}
\and
\IEEEauthorblockN{Thomas Hiessl}
\IEEEauthorblockA{Siemens Technology}
\and
\IEEEauthorblockN{Safoura Rezapour Lakani}
\IEEEauthorblockA{Siemens Technology}
\and
\IEEEauthorblockN{Daniel Schall}
\IEEEauthorblockA{Siemens Technology}
\and
\IEEEauthorblockN{Clemens Heitzinger}
\IEEEauthorblockA{TU Wien}
\and
\IEEEauthorblockN{Jana Kemnitz}
\IEEEauthorblockA{Siemens Technology}
}

\maketitle

\begin{abstract}
Federated Learning (FL) decouples model training from the need for direct access to the data and allows organizations to collaborate with industry partners to reach a satisfying level of performance without sharing vulnerable business information. The performance of a machine learning algorithm is highly sensitive to the choice of its hyperparameters. In an FL setting, hyperparameter optimization poses new challenges. In this work, we investigated the impact of different hyperparameter optimization approaches in an FL system. In an effort to reduce communication costs, a critical bottleneck in FL, we investigated a local hyperparameter optimization approach that -- in contrast to a global hyperparameter optimization approach -- allows every client to have its own hyperparameter configuration. We implemented these approaches based on grid search and Bayesian optimization and evaluated the algorithms on the MNIST data set using an i.i.d.\ partition and on an Internet of Things (IoT) sensor based industrial data set using a non-i.i.d.\ partition. 
\end{abstract}


\section{Introduction}

The performance of a machine learning algorithm is highly sensitive to the choice of its hyperparameters. Therefore, hyperparameter selection is a crucial task in the optimization of knowledge-aggregation algorithms. Federated Learning (FL) is a recent machine learning approach which aggregates machine learning model parameters between devices (henceforth clients) without sharing their data. The aggregation is coordinated by a server. Industrial Federated Learning (IFL) is a modified approach of FL in an industrial context \cite{c3}. In an FL setting, hyperparameter optimization poses new challenges and is a major open research area \cite{c6}. In this work, we investigate the impact of different hyperparameter optimization approaches in an IFL system. We believe that the data distribution influences the choice of the best hyperparameter configuration and suggest that the best hyperparameter configuration for a client might differ from another client based on individual data properties. Therefore, we investigate a local hyperparameter optimization approach that -- in contrast to a global hyperparameter optimization approach -- allows every client to have its own hyperparameter configuration. The local approach allows us to optimize hyperparameters prior to the federation process reducing communication costs.

Communication is considered a critical bottleneck in FL \cite{c10}. Clients are usually limited in terms of communication bandwidth enhancing the importance of reducing the number of communication rounds or using compressed communication schemes for the model updates to the central server \cite{c10}. Dai et al.\ \cite{c5} introduced \textit{Federated Bayesian Optimization} (FBO) extending Bayesian optimization to the FL setting. However, until now, there is no research on the impact of global and local hyperparameter optimization in FL. Therefore, we compare a local hyperparameter optimization approach to a global hyperparameter optimization approach, optimizing hyperparameters in the federation process.

The aim of this work is to i) analyze challenges and formal requirements in FL, and in particular in IFL, ii) to evaluate the performance of an Internet of Things (IoT) sensor based classification task in an IFL system, iii) to investigate a communication efficient hyperparameter optimization approach, and iv) to compare different hyperparameter optimization algorithms. Therefore, we want to answer the following questions. 
\begin{itemize}
    \item[Q1:] Does FL work for an IoT sensor based anomaly classification task on industrial assets with non-identically distributed data in an IFL system with a cohort strategy?
	\item[Q2:] Can we assume that the global and local hyperparameter optimization approach deliver the same hyperparameter configuration in an i.i.d.\ FL setting?
	\item[Q3:] Can we reduce communication costs in the hyperparameter optimization of a non-i.i.d.\ classification task in context of FL by optimizing a hyperparameter locally prior to the federation process?
	\item[Q4:] Does Bayesian optimization outperform grid search, both in a global and local approach of a non-i.i.d.\ IoT sensor based classification task?
\end{itemize}

\section{Algorithmic Challenges and Formal Requirements for industrial Assets}

In FL, new algorithmic challenges arise that differentiate the corresponding optimization problem from a distributed optimization problem. In distributed learning settings, major assumptions regarding the training data are made which usually fail to hold in an FL setting \cite{c1}. Moreover, non-i.i.d.\ data, limited communication, and limited and unreliable client availability pose further challenges for optimization problems in FL \cite{c6}. Kairouz et al.\ \cite{c6} considered the need for addressing these challenges as a major difference to distributed optimization problems. The optimization problem in FL is therefore referred to as federated optimization emphasizing the difference to distributed optimization \cite{c1}. In an IFL setting, additional challenges regarding industrial aspects arise \cite{c3}. In this section, we want to formulate the federated optimization problem and discuss the algorithmic challenges of FL in general, and in particular of IFL.

\subsection{Problem Formulation}

We consider a supervised learning task with features $x$ in a sample space $\mathcal{X}$ and labels $y$ in a label space $\mathcal{Y}$. We assume that we have $K$ available clients, $K \in \mathbb{N}_{\ge2}$, with 
\begin{displaymath}
D_{k}:=D_{\mathcal{X},k}\times D_{\mathcal{Y},k} \subseteq \mathcal{X} \times \mathcal{Y}
\end{displaymath}
denoting the data set of client $k$ and $n_{k}:=|D_{k}|$ denoting the cardinality of the client's data set. Let $\mathcal{Q}$ denote the distribution over all clients, and let $\mathcal{P}_{k}$ denote the data distribution of client $k$. We can then access a specific data point by first sampling a client $k\sim \mathcal{Q}$ and then sampling a data point $(x,y) \sim \mathcal{P}_{k}$ \cite{c6}. Then, the local objective function is 
\begin{align}
F_{k}(w):=\underset{(x,y) \sim \mathcal{P}_{k}}{\mathbb{E}}[f(x,y,w)],
\end{align}
where $w \in \mathbb{R}^{d}$ represents the parameters of the machine learning model and $f(x,y,w)$ represents the loss of the prediction on sample $(x,y)$ for the given parameters $w$. Typically, we wish to minimize
\begin{align}
F(w):=\frac{1}{K} \sum_{k=1}^{K} F_{k}(w).  
\end{align}

\subsection{Federated Learning}

One of the major challenges concerns data heterogeneity. In general, we cannot assume that the data is identically distributed over the clients, that is $\mathcal{P}_k = \mathcal{P}_l$ for all~$k$ and~$l$. Therefore, $F_{k}$ might be an arbitrarily bad approximation of~$F$ \cite{c1}. 

In the following, we want to analyze different non-identically distributed settings as demonstrated by Hsieh et al.\ \cite{c8} assuming that we have an IoT sensor based anomaly classification task in an industrial context. Given the distribution $\mathcal{P}_{k}$, let $P^{k}_{\mathcal{X},\mathcal{Y}}$ denote the bivariate probability function, let $P^{k}_{\mathcal{X}}$ and $P^{k}_{\mathcal{Y}}$ denote the marginal probability function respectively. Using the conditional probability function $P^{k}_{\mathcal{Y}|\mathcal{X}}$ and $P^{k}_{\mathcal{X}|\mathcal{Y}}$, we can now rewrite the bivariate probability function as
\begin{align}
P^{k}_{\mathcal{X},\mathcal{Y}}(x,y) = P^{k}_{\mathcal{Y}|\mathcal{X}}(y|x) P^{k}_{\mathcal{X}}(x)=P^{k}_{\mathcal{X}|\mathcal{Y}}(x|y) P^{k}_{\mathcal{Y}}(y)
\end{align} 
for $(x,y) \in \mathcal{X} \times \mathcal{Y}$. This allows us to characterize different settings of non-identically distributed data:

\textit{Feature distribution skew:}
We assume that $P^{k}_{\mathcal{Y}|\mathcal{X}}=P^{l}_{\mathcal{Y}|\mathcal{X}}$ for all $k$, $l$, but $P^{k}_{\mathcal{X}} \ne P^{l}_{\mathcal{X}}$ for some $k$, $l$. Clients that have the same anomaly classes might still have differences in the measurements due to variations in sensor and machine type.

\textit{Label distribution skew:} 
We assume that $P^{k}_{\mathcal{X}|\mathcal{Y}}=P^{l}_{\mathcal{X}|\mathcal{Y}}$ for all $k$, $l$, but $P^{k}_{\mathcal{Y}}\ne P^{l}_{\mathcal{Y}}$ for some $k$, $l$. The distribution of labels might vary across clients as clients might experience different anomaly classes. 

\textit{Same label, different features:} 
We assume that $P^{k}_{\mathcal{Y}}=P^{l}_{\mathcal{Y}}$ for all $k$, $l$, but $P^{k}_{\mathcal{X}|\mathcal{Y}}\ne P^{l}_{\mathcal{X}|\mathcal{Y}}$ for some $k$, $l$. The same anomaly class can have significantly different features for different clients due to variations in machine type, operational- and environmental conditions. 

\textit{Same features, different label:}
We assume that $P^{k}_{\mathcal{X}}=P^{l}_{\mathcal{X}}$ for all $k$ and $l$, but $P^{k}_{\mathcal{Y}|\mathcal{X}}\ne P^{l}_{\mathcal{Y}|\mathcal{X}}$ for some $k$, $l$. The same features can have different labels due to operational- and environmental conditions, variation in manufacturing, maintenance et cetera.

\textit{Quantity skew:} 
We cannot assume that different clients hold the same amount of data, that is $n_{k} = n_{l}$ for all $k$, $l$. Some clients will generate more data than others.

In real-world problems, we expect to find a mixture of these non-identically distributed settings. In FL, heterogeneity does not exclusively refer to a non-identical data distribution, but also addresses violations of independence assumptions on the distribution $\mathcal{Q}$ \cite{c6}. Due to limited, slow and unreliable communication on a client, the availability of a client is not guaranteed for all communication rounds. Communication is considered a critical bottleneck in FL \cite{c10}. In each communication round, the participating clients send a full model update $w$ back to the central server for aggregation. In a typical FL setting, however, the clients are usually limited in terms of communication bandwidth \cite{c10}. Consequently, it is crucial to minimize the communication costs.

\subsection{Industrial Federated Learning}

In an industrial setting, FL experiences challenges that specifically occur in an industrial context. Industrial assets have access to a wealth of data suitable for machine learning models, however, the data on an individual asset is typically limited and private in nature. In addition to sharing the model within the company, it can also be shared with an external industry partner \cite{c3}. FL leaves possibly critical business information distributed on the individual client (or within the company). However, Zhao et al.\ \cite{c4} proved that heterogeneity, in particular, a highly skewed label distribution, significantly reduces the accuracy of the aggregated model in FL. In an industrial context, we expect to find heterogeneous clients due to varying environmental and operational conditions on different assets. Therefore, Hiessl et al.\ \cite{c3} introduced a modified approach of FL in an industrial context and termed it \textit{Industrial Federated Learning} (IFL). IFL does not allow arbitrary knowledge exchange between clients. Instead, the knowledge exchange only takes place between clients that have sufficiently similar data. Hiessl et al.\ \cite{c3} refer to this set of clients as a \textit{cohort}. We expect the federated learning approach in a cohort to approximate the corresponding central learning approach.

\section{Hyperparameter Optimization Approaches in an IFL System}

In an FL setting, hyperparameter optimization poses new challenges and is a major open research area \cite{c6}. The performance of a machine learning model is linked to the amount of communication \cite{c9}. In an effort to reduce communication costs, a critical bottleneck in FL \cite{c10}, we investigated a communication efficient hyperparameter optimization approach, a local hyperparameter optimization approach that allows us to optimize hyperparameters prior to the federation process. Kairouz et al.\ \cite{c6} introduced the idea of a separate optimization of hyperparameters and suggest a different hyperparameter choice for dealing with non-i.i.d.\ data. 

Dai et al.\ \cite{c5} investigated a communication efficient local hyperparameter optimization approach and introduced Federated Bayesian Optimization (FBO) extending Bayesian optimization to the FL setting. In FBO, every client locally uses Bayesian optimization to find the optimal hyperparameter configuration. Additionally, each client is allowed to request for information from other clients. Dai et al. \cite{c5} proved a convergence guarantee for this algorithm and its robustness against heterogeneity. However, until now, there is no research on the impact of global and local hyperparameter optimization.

In the LocalHPO algorithm \ref{alg: LocalHPO}, we perform local hyperparameter optimization. We optimize the hyperparameter configuration $\lambda^{k}$ for each client $k$. In the GlobalHPO algorithm \ref{alg: GlobalHPO}, we perform global hyperparameter optimization. We optimize the hyperparameter configuration $\lambda$ in the federation process. The LocalOptimization method in the LocalHPO algorithm \ref{alg: LocalHPO} and the GlobalOptimization method in the GlobalHPO algorithm \ref{alg: GlobalHPO} can be based on any hyperparameter optimization algorithm.
\begin{algorithm}\small
	\DontPrintSemicolon
	\caption{LocalHPO}
	\label{alg: LocalHPO}
	\SetAlgoLined
	\textbf{Server executes:}\;
	initialize $w_{0}$\;
	\For{each client $k=1,\dots,K$}{
		$\lambda^{k} :=$ LocalOptimization$(k, w_{0})$\;
	}
	return $(\lambda^{k})_{k=1}^{K}$
\end{algorithm}
\begin{algorithm}\small
	\DontPrintSemicolon
	\caption{GlobalHPO}
	\label{alg: GlobalHPO}
	\SetAlgoLined
	\textbf{Server executes:}\;
	$\lambda :=$ GlobalOptimization$()$\;
	return $\lambda$
\end{algorithm}

We want to differentiate between a global hyperparameter $\lambda_{i}$ whose value is constant for all clients and a local hyperparameter $\lambda_{i}^{k}$ whose value depends on a client $k$. Here, $\lambda_{i}^{k}$ denotes the hyperparameter $\lambda_{i}$ on client $k$. We notice that this differentiation is only relevant for settings with non-i.i.d.\ data. In an i.i.d.\ setting, we assume that a hyperparameter configuration that works for one client also works for another client. In our experiments, we verified this assumption for a proxy data set. 

\section{Data, Algorithms and Experiments}

In the next section, we want to make our benchmark design explicit and present our experimental setup. We will present the machine learning tasks including the data partition of the training data, the machine learning models, the optimization algorithms and our experiments. We considered an image classification task on a data set, the MNIST data set of handwritten digits, and an IoT sensor based anomaly classification task on industrial assets. 

\subsection{Data}

In order to test the IFL system on the MNIST data set, we still need to specify on how to distribute the data over artificially designed clients. To systematically evaluate the effectiveness of the IFL system, we simulated an i.i.d.\ data distribution. This refers to shuffling the data and partitioning the data into $10$ clients, each receiving 6\,000 examples. Following the approach of McMahan et al.\ \cite{c1}, we applied a convolutional neural network with the following settings: $2$~convolutional layers with $32$ and $64$ filters of size $5\!\times\!5$ and a ReLu activation function, each followed by a max pooling layer of size $2\!\times\!2$, a dense layer with $512$ neurons and a ReLu activation function, a dense layer with $10$ neurons and a softmax activation function.

The industrial task concerns IoT sensor based anomaly classification on industrial assets. The data was acquired with the SITRANS multi sensor, specifically developed for industrial applications and its requirements \cite{c11}. We considered multiple centrifugal pumps with sensors placed at different positions, in different directions to record three axis vibrational data in a frequency of $6644$\,Hz. Per minute, $512$ samples were collected. We operated the pumps under $6$~varying conditions, including $3$~healthy states and $3$~anomalous states. A client is either assigned data of an asset in a measurement, or data of several assets in a measurement ensuring that each client sees all operating conditions. However, since in the process of measurement, the assets were completely dismantled and rebuilt, we consider the data to be non-i.i.d.\ regarding its feature distribution. We applied an artificial neural network with the following settings: a dense layer with $64$ neurons and a ReLu activation function, a dropout layer with a dropout rate of $0.4$, a dense layer with $6$ neurons and a ReLu activation function, a dropout layer with a dropout rate of $0.4$, and a softmax activation function. We remapped the features using the Kabsch algorithm \cite{c12}, applied a sliding window, extracted the Melfrequency cepstral coefficients, applied the synthetic minority oversampling technique \cite{c12}, and normalized the resulting features.

\subsection{Algorithms}

Our evaluations include the Federated Averaging (FedAvg) algorithm according to McMahan et al.\ \cite{c1}, and the hyperparameter optimization approaches LocalHPO \ref{alg: LocalHPO} and GlobalHPO \ref{alg: GlobalHPO}. We implemented these approaches based on grid search and Bayesian optimization. In this section, we give their pseudocode. We searched for the learning rate $\eta$ with fixed fraction of participating clients $C$, number of communication rounds $R$, number of local epochs $E$, and mini-batch size $B$.

In algorithm \ref{alg: LocalOptimization Grid}, we give the pseudocode of the LocalOptimization method in LocalHPO \ref{alg: LocalHPO} based on the grid search algorithm with a fixed grid $G$. We iterate through the grid~$G$, train the model on the training data of client $k$ based on the ClientUpdate method used in the FedAvg algorithm \cite{c1} with the learning rate $\eta$ as an additional argument, and validate the performance of the model $w_{\eta}$ on the validation data $\mathcal{D}_{\mathrm{valid}}^{k}$ of client $k$. Finally, the learning rate that yields the highest accuracy $A_{\eta}$ on the validation data is selected. Here, $w_{\eta}$ denotes the resulting model trained on the training data with learning rate $\eta$ and $A(\mathcal{D}_{\mathrm{valid}}^{k}, w_{\eta})$ denotes the accuracy of the model tested on the validation data $\mathcal{D}_{\mathrm{valid}}^{k}$ of client $k$.

\begin{algorithm}\small
	\DontPrintSemicolon
	\caption{Local Grid Search}
	\label{alg: LocalOptimization Grid}
	\SetAlgoLined
	LocalOptimization$(k, w_{0})$:\;
	\For{\text{each learning rate} $ \eta \in G$}{
		$w_{\eta} :=$ ClientUpdate$(k, w_{0}, \eta)$\;
		$A_{\eta} := A(\mathcal{D}_{\mathrm{valid}}^{k}, w_{\eta})$
	}
	$\eta_{k}^{*} := \underset{\eta \in G}{\arg \max} \hspace{1mm} A_{\eta}$\;
	return $\eta_{k}^{*}$
\end{algorithm}
\begin{algorithm}\small
	\DontPrintSemicolon
	\caption{Global Grid Search}
	\label{alg: GlobalOptimization Grid}
	\SetAlgoLined
	GlobalOptimization$()$:\;
	\For{\text{each learning rate} $ \eta \in G$}{
		$w_{\eta} :=$ FederatedAveraging$(\eta)$\;
		\For{\text{each client} $k=1,\dots,K$}{
			$A_{\eta}^{k} := A(\mathcal{D}_{\mathrm{valid}}^{k}, w_{\eta})$
		}
		$A_{\eta} := \frac{1}{K}\sum_{k=1}^{K}A_{\eta}^{k}$
	}
	$\eta^{*} := \underset{\eta \in G}{\arg \max} \hspace{1mm} A_{\eta}$\;
	return $\eta^{*}$
\end{algorithm}
In algorithm \ref{alg: GlobalOptimization Grid}, we give the pseudocode of the GlobalOptimization method in GlobalHPO \ref{alg: GlobalHPO} based on the grid search algorithm with a fixed grid $G$. We iterate through the grid, perform the FedAvg algorithm \cite{c1} with the learning rate $\eta$ as an additional argument, validate the performance of the model $w_{\eta}$ on the validation data $\mathcal{D}_{\mathrm{valid}}^{k}$ for all clients $k$ and compute the average accuracy of all clients. Finally, the learning rate that yields the highest average accuracy $A_{\eta}$ is selected. 

In algorithm \ref{alg: LocalOptimization Bayesian}, we give the pseudocode of the LocalOptimization method in LocalHPO \ref{alg: LocalHPO} based on Bayesian optimization. The objective function $f$ takes the learning rate $\eta$ as an argument, trains the model on the training data of client $k$ based on the ClientUpdate method used in the FedAvg algorithm \cite{c1} with the learning rate $\eta$ as an additional argument, validates the performance of the model $w$ on the validation data $\mathcal{D}_{\mathrm{valid}}^{k}$ of client $k$, and returns the resulting accuracy. We initialize a gaussian process $GP$ for the objective function $f$ with $n_{\mathrm{init}}$ sample points. Then, we find the next sample point $\eta_{n_{\mathrm{init}} + i}$ by maximizing the acquisition function, evaluate $f(\eta_{n_{\mathrm{init}} + i})$, and update the gaussian process $GP$. Finally, we select the learning rate $\eta^{*}$ that yields the highest accuracy. We repeat this for $n_{\mathrm{iter}}$ iterations.

In algorithm \ref{alg: GlobalOptimization Bayesian}, we give the pseudocode of the GlobalOptimization method in GlobalHPO \ref{alg: GlobalHPO} based on Bayesian optimization. The objective function $f$ takes the learning rate $\eta$ as an argument, performs the FedAvg algorithm \cite{c1} with the learning rate $\eta$ as an additional argument, validates the performance of the model $w$ on the validation data $\mathcal{D}_{\mathrm{valid}}^{k}$ for all clients $k$, computes the average accuracy of all clients and returns the resulting accuracy. We initialize a gaussian process $GP$ for the objective function $f$ with $n_{\mathrm{init}}$ sample points. Then, we find the next sample point $\eta_{n_{\mathrm{init}} + i}$ by maximizing the acquisition function, evaluate $f(\eta_{n_{\mathrm{init}} + i})$, and update the gaussian process $GP$. Finally, we select the learning rate $\eta^{*}$ that yields the highest average accuracy. We repeat this for $n_{\mathrm{iter}}$ iterations.

\begin{algorithm}\small
	\DontPrintSemicolon
	\caption{Local Bayesian Optimization}
	\label{alg: LocalOptimization Bayesian}
	\SetAlgoLined
	LocalOptimization$(k, w_{0})$:\;
	initialize a gaussian process $GP$ for $f$\;
	evaluate $f$ at $n_{\mathrm{init}}$ initial points\;
	\For{$i=1,\dots,n_{\mathrm{iter}}$}{
		find sample point $\eta_{n_{\mathrm{init}} + i}$ that maximizes acquisition function\;
		evaluate objective function $f$ at $\eta_{n_{\mathrm{init}} + i}$\; 
		update the gaussian process $GP$\;
	}
	$\eta^{*} := \underset{i=1, \dots, n_{\mathrm{init}}+n_{\mathrm{iter}}}{\arg \max} \hspace{1mm} f(\eta_{i})$\;
	return $\eta^{*}$\;
	
	\textbf{objective function:}\;
	$f(\eta)$:\;
	$w :=$ ClientUpdate$(k, w_{0}, \eta)$\;
	$A := A(\mathcal{D}_{\mathrm{valid}}^{k}, w)$\;
	return $A$
\end{algorithm}

\begin{algorithm}\small
	\DontPrintSemicolon
	\caption{Global Bayesian Optimization}
	\label{alg: GlobalOptimization Bayesian}
	\SetAlgoLined
	GlobalOptimization$()$:\;
	initialize a gaussian process $GP$ for $f$\;
	evaluate $f$ at $n_{\mathrm{init}}$ initial points\;
	\For{$i=1,\dots,n_{\mathrm{iter}}$}{
		find sample point $\eta_{n_{\mathrm{init}} + i}$ that maximizes acquisition function\;
		evaluate objective function $f$ at $\eta_{n_{\mathrm{init}} + i}$\; 
		update the gaussian process $GP$\;
	}
	$\eta^{*} := \underset{i=1,\dots, n_{\mathrm{init}}+n_{\mathrm{iter}}}{\arg \max} \hspace{1mm} f(\eta_{i})$\;
	return $\eta^{*}$\;
	
	\textbf{objective function:}\;
	$f(\eta)$:\;
	$w :=$ FederatedAveraging$(\eta)$\;
	\For{\text{each client} $k=1,\dots,K$}{
		$A^{k} := A(\mathcal{D}_{\mathrm{valid}}^{k}, w)$\;
	}
	$A := \frac{1}{K}\sum_{k=1}^{K}A^{k}$\;
	return A
\end{algorithm}

\subsection{Experiments}

In order to systematically investigate the impact of global and local hyperparameter optimization, we compared the global and local hyperparameter optimization approach in an i.i.d.\ setting, the MNIST machine learning task, as well as in a non-i.i.d.\ setting, the industrial task. Therefore, we implemented the global and local optimization approach based on grid search with a grid $G:=[0.0001, 0.001, 0.01, 0.1]$, and based on Bayesian optimization with the widely used squared exponential kernel and the upper confidence bound acquisition function. We searched for the learning rate $\eta$ with fixed $R$, $C$, $E$ and $B$. 

In order to evaluate the global and local optimization approaches in a direct comparison, we chose the number of epochs $E$ in the local optimization approach as $E=E_{\mathrm{global}}R$, where $E_{\mathrm{global}}$ is the number of epochs in the global optimization approach and $R$ is the number of communication rounds. In the global optimization task, we set $R:=10$, $C:=1$, $E:=1$ and $B:=128$ for the MNIST data, and $R:=10$, $C:=1$, $E:=5$ and $B:=128$ for the industrial data. In the local optimization task, we set $E:=10$ and $B:=128$ for the MNIST data, and $E:=50$ and $B:=128$ for the industrial data. For the evaluation of the global hyperparameter optimization approach, we optimized the learning rate using the global approach, trained the federated model with a global learning rate, and tested the resulting federated model on the cohort test data. Then, we optimized the learning rate using the local approach, trained the federated model with local individual learning rates for each client in the cohort, and tested the resulting federated model on the cohort test data.

\section{Experimental Results}
\begin{figure}[ht]
	\centerline{\includegraphics[width=0.35\textwidth]{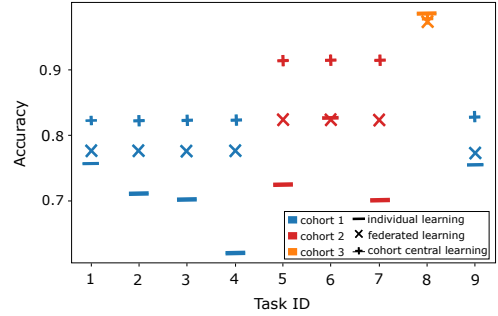}}
	\caption{Comparison of individual learning, central learning, and federated learning on the industrial data set.}
	\label{client comparison sp260 grid}
\end{figure}

Following the approach of Hiessl et al.\ \cite{c3}, we demonstrated the effectiveness of the IFL System for the industrial task and showed that the IFL approach performs better than the individual learning approach and approximates the central learning approach. Fig.~\ref{client comparison sp260 grid} shows the test accuracy on the central cohort test data for each client, for i) a model trained on the individual training data of the client, ii) a central model trained on the collected training data of all clients in the cohort, and iii) the federated model trained in the cohort.

\begin{figure}[ht]
	\centerline{\includegraphics[width=0.4\textwidth]{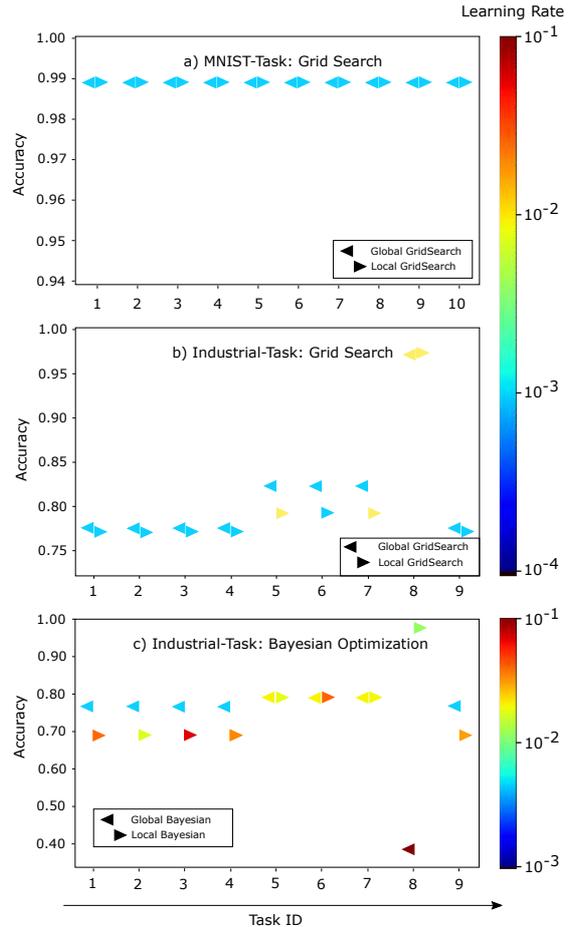}}
	\caption{Comparison of the optimization approaches based on a) grid search for the MNIST task, b) grid search for the industrial task, and c) Bayesian optimization for the industrial task.}
	\label{hpo comparison}
\end{figure}

Fig.\ \ref{hpo comparison} a) shows the results for the MNIST data. The optimization approaches are based on the grid search algorithm. For the training posterior to the optimization, we set $R:=10$, $C:=1$, $E:=1$, and $B:=128$ in the IFL system. The color indicates the optimized learning rate on the corresponding client. Since the MNIST data is i.i.d., there is only one cohort and all clients have the same federated model and thus the same test accuracy. Our results show that the grid search algorithm selected $10^{-3}$  in the local optimization of the learning rate on each client. According to our expectation, the global optimization approach yielded the same learning rate.

For the industrial task, we evaluated the global and local optimization approach based on grid search and Bayesian optimization. For the training posterior to the optimization, we set $R:=20$, $C:=1$, $E:=5$, and $B:=128$ in the IFL system. Fig.~\ref{hpo comparison}~b) shows the results for the industrial data with the optimization approaches based on the grid search algorithm. The results show that, in all cohorts, the global approach yielded an equal or larger accuracy than the local approach.

Fig.\ \ref{hpo comparison} c) shows the results for the industrial data with the optimization approaches based on the Bayesian algorithm. Note that the search space of the learning rate was $[10^{-4}, 10^{-1}]$ in the optimization while the scale in the plot starts from $10^{-3}$. The results show that the global approach yielded a larger accuracy than the local approach in cohort~$0$ and cohort~$1$. 

The local Bayesian approach yielded different learning rates, see Fig.~\ref{hpo comparison}~c), on clients with no difference in data, that is, the same number of samples, the same class distribution, and the same measurement protocol. However, the local grid search approach yielded the same learning rate as the global grid search approach, see Fig.~\ref{hpo comparison}~b). Therefore, we suggest that the reason lies in the implementation of the Bayesian optimization approach and a not sufficiently large number of iterations to guarantee convergence. 

In order to compare the optimization approaches for the industrial task, we performed a paired t-test regarding the test accuracy to determine the statistical significance, see table~\ref{tab2}. We observe that the global optimization approach is significantly better than the local approach, both for the grid search approach ($p=0.028$) and for the Bayesian approach ($p=0.012$). Furthermore, the results show that the grid search approach is significantly better than the Bayesian approach, both for the global approach ($p=0.004$) and for the local approach ($p=0.008$). Note that we considered cohort~$2$ an outlier and excluded this cohort from our calculations. Cohort~$2$ only consists of client~$8$, a client whose data was not generated according to the standard measurement protocol. Without outlier removal, the global grid search approach is still significantly better than the local grid search approach ($p=0.032$), and the local grid search approach is significantly better than the local Bayesian approach ($p=0.010$). However, there is no significant difference in the global Bayesian approach vs.\ the local Bayesian approach ($p=0.755$) and in the global grid search approach vs.\ the global Bayesian approach ($p=0.230$).

\section{Conclusion and Future Work}

The results show that the federated learning approach approximates the central learning approach, while outperforming individual learning of the clients. In this work, we investigated the impact of global and local optimization approaches in an IFL System based on a proxy data set and a real-world problem. In our experiments on the industrial data, local optimization yielded different learning rates on different clients in a cohort. However, the results show that a globally optimized learning rate, and thus, a global learning rate for all clients in a cohort improves the performance of the resulting federated model. Therefore, we conclude that the global optimization approach outperforms the local optimization approach resulting in a communication-performance trade-off in the hyperparameter optimization in FL. In our experiments on the proxy data set, however, the local approach achieved the same performance as the global approach. 

\begin{table}[t]
	\caption{Test accuracy of federated model on central cohort test data posterior to corresponding optimization approach and training}
	\begin{center}
		\begin{tabular}{ |c|c|c|c|c| } 
			\hline
		    client & global grid & local grid & global Bayesian & local Bayesian \\ 
		    \hline
	    	$1$ & \textbf{0.7756} & 0.7720 &  0.7659 & 0.6897 \\ 
		    \hline
		    $2$ & \textbf{0.7756} & 0.7720 &  0.7659 & 0.6897 \\ 
		    \hline
		    $3$ & \textbf{0.7756} & 0.7720 &  0.7659 & 0.6897 \\ 
		    \hline
		    $4$ & \textbf{0.7756} & 0.7720 &  0.7659 & 0.6897 \\  
		    \hline
		    $5$ & \textbf{0.8230} & 0.7921 & 0.7882 & 0.7889 \\ 
		    \hline
		    $6$ & \textbf{0.8230} & 0.7921 & 0.7882 & 0.7889 \\  
		    \hline
		    $7$ & \textbf{0.8230} & 0.7921 & 0.7882 & 0.7889 \\ 
		    \hline
		    $8$ &  0.9740 & \textbf{0.9749} &  0.3867 & 0.9736\\
		    \hline
		    $9$ &  \textbf{0.7756} & 0.7720 &  0.7659 & 0.6897 \\ 
		    \hline
	    \end{tabular}
		\label{tab2}
	\end{center}
\end{table}

A limitation of our study is that we only considered one hyperparameter in our optimization task. Hence it would be interesting to explore whether we can confirm these observations for a hyperparameter configuration of more hyperparameters. The results show that the grid search approaches outperform the Bayesian approaches, both globally and locally. However, we suggest a convergence analysis for the Bayesian approach. 

\nocite{*}
\bibliography{conference_idsc}{}
\bibliographystyle{IEEEtran}

\end{document}